\def\year{2021}\relax
\newif\ifsubmission
\documentclass[letterpaper]{article} \usepackage{aaai21_preprint}
\nocopyright \pdfoutput=1
 \usepackage{times}  \usepackage{helvet} \usepackage{courier}  \usepackage[hyphens]{url}  \usepackage{appendix}
\usepackage{graphicx}    \usepackage{natbib}  \usepackage{caption}       \usepackage{xcolor} \usepackage{graphicx}
\usepackage{amsmath}
\usepackage{amssymb}
\usepackage[ruled]{algorithm2e}

\title{Relative Variational Intrinsic Control}
\author{
	\ifsubmission
	Anonymous Authors
	\else
Kate Baumli,
		David Warde-Farley,
		Steven Hansen,
		Volodymyr Mnih \\
		\fi
}
\affiliations{
	\ifsubmission
	Anonymous Affiliations
	\else
    DeepMind \\
\{baumli, dwf, stevenhansen, vmnih\} @google.com
    \fi
}
\begin{document}

\maketitle
\begin{abstract}
In the absence of external rewards, agents can still learn useful behaviors by identifying and mastering a set of diverse skills within their environment.
Existing skill learning methods use mutual information objectives to incentivize each skill to be diverse and distinguishable from the rest.
However, if care is not taken to constrain the ways in which the skills are diverse, trivially diverse skill sets can arise.
To ensure useful skill diversity, we propose a novel skill learning objective, Relative Variational Intrinsic Control (RVIC), which incentivizes learning skills that are distinguishable in how they change the agent's relationship to its environment.
The resulting set of skills tiles the space of affordances available to the agent. 
We qualitatively analyze skill behaviors on multiple environments and show how RVIC skills are more useful than skills discovered by existing methods when used in hierarchical reinforcement learning.
 \end{abstract}

\section{Introduction}
Deep reinforcement learning (RL) methods have demonstrated the ability to successfully learn to achieve a task defined by a reward function in a variety of domains \cite{sutton1998book, mnih2015human, silver2017chess, silver2017mastering}.
However, the knowledge obtained by an RL agent is usually highly specific to the particular task it was trained on, and is not well suited towards transfer or generalization \cite{whiteson2011protecting, cobbe2019quantifying}.

In contrast, humans can obtain and maintain repurposable knowledge about their environments and how they can behave in them, even in the absence of an explicit end goal or reward.
We maintain sets of skills that can transfer from one task to the next.
For example, we can learn a general skill to throw an object and we can apply slightly modified versions of that skill in different contexts to enable us to throw a paper airplane, a baseball, or a water balloon.

Mutual information based skill discovery methods such as Variational Intrinsic Control (VIC) \cite{gregor16variational} and Diversity Is All You Need (DIAYN) \cite{eysenbach2018diversity} offer a promising direction for increasing practical applicability of deep reinforcement learning methods to real-world problems.
By first learning useful skills purely from intrinsic rewards, agents can repurpose the learned skills to solve more challenging downstream tasks specified by extrinsic rewards.

These methods rely on the idea of inverse predictability for skill learning, i.e. that it should be possible to infer the skill used to generate a trajectory from the states in the trajectory. This requires each skill to be distinguishable from the others, ensuring a diverse set of skills. 
The conditioning of the inverse predictor varies from relying on any state along the trajectory in DIAYN to using the whole trajectory in VALOR \cite{achiam2018variational}.
We focus on a common intermediate case, introduced by VIC, where the inverse predictor relies on the first and last states in the trajectory.

In practice, these inverse predictability objectives are often trivial enough to achieve using only information from the end of the trajectory that they ignore given information about the beginning of the trajectory.
This results in a set of skills that simply partitions the state space based on
where the agent ends up at the end of each skill.
We argue that this behavior is undesirable because it limits the usefulness (in terms of transferability and generalizability) of the skill set.
A set of skills which each correspond to a specific target state (or a small cluster of nearby target  states) is limited to only ever going between those specific regions of the state space. 

In this paper, we propose a way of acquiring more composable and generalizable skills, making note that a skill's behavior should differ depending on the
current state of the agent. 
For example, the skills an agent is afforded while sitting on an airplane with a fastened seatbelt are different than the skills afforded on a spacious football field.
However, skills should generalize between different areas of the state space; picking up a dropped pen on the airplane and picking up a football are in a sense the same skill, but performing the "picking up" skill doesn't arrive at the same target state in both cases.
After performing the skill, the agent's new state should be on the plane with a pen in hand and on the field with a football in hand, respectively. 
In other words, the final state that the skill arrives at should depend on both the skill itself and the agent's initial state. 

To incentivize more meaningfully diverse skill sets, we propose a new skill learning method, Relative Variational Intrinsic Control (Relative VIC / RVIC) that utilizes two inverse predictors: one which relies on the first and last states of the trajectory and one that only relies on the last state to predict the skill.
Incentivizing the agent to maximize predictability with respect to the former and minimize predictability with respect to the latter, skills are forced to be relative to the agent's state at the beginning of the skill, guarding against state space partitioning. Instead, RVIC skills partition the space of affordances \cite{gibson1977theory}-- that is, they are diverse in the way that they change the agent's relationship to its environment. 

In this paper, we introduce the Relative VIC skill learning method, qualitatively examine the skills on several domains, and show that they are more useful in a hierarchical RL set up than skills learned by existing methods.

 \section{Background}
\label{sec:methods}
We consider a skill-conditional policy $\pi_\theta(\cdot; \Omega)$ which maps states $s$ to a distribution over actions $a$.
In this work, we will assume that $\Omega$ is discrete although the algorithms can also be applied to continuous skills.
We will train this policy on episodes wherein we sample a skill $\Omega$ uniformly from the set of available skills and follow it for a fixed number of steps, $T$.
We refer to the states $s_0, \ldots, s_T$ and corresponding actions $a_1, \ldots, a_T$ as a skill trajectory or a skill episode.

\citet{gregor16variational} introduced the idea of discovering skills by maximizing the mutual information between $\Omega$ and $s_T$, the final state of a skill trajectory, conditioned on the first state $s_0$.
Their approach, known as Variational Intrinsic Control (VIC), relies on the well known lower bound of \citet{barber2004im} on the mutual information.
When applied to the VIC objective, the lower bound takes the form
\begin{align}
	I(s_T, \Omega|s_0)  &= H(\Omega|s_0) - H(\Omega|s_T,s_0) \nonumber \\
	&\geq  H(\Omega|s_0) + \mathbb{E}_\Omega\mathbb{E}_{s_0, s_T\sim\pi_\Omega} \log q_\phi(\Omega|s_T, s_0)
\end{align}
where $q$ is a variational distribution.
Following \citet{eysenbach2018diversity}, we assume that skills $\Omega$ are sampled from a fixed distribution.
Optimizing this objective involves two separate optimization steps.
The first performs gradient ascent on $\log q_\phi(\Omega|s_T, s_0)$ with respect to variational parameters $\phi$.
This corresponds to training $q$ to be an inverse predictor which can infer the skill $\Omega$ used to generate the skill episode from the first and final states.
The second optimization step involves optimizing the parameters of the skill-conditional policy $\pi_\theta(\cdot; \Omega)$ using a reinforcement learning algorithm with a sparse reward that is proportional to $\log q_\phi(\Omega|s_T, s_0)$ at time $T$ and zero for all other time steps\footnote{Note that we omit the constant additive term that derives from the entropy of the skill prior.}.

\begin{figure}[t]
\includegraphics[width=0.46\textwidth]{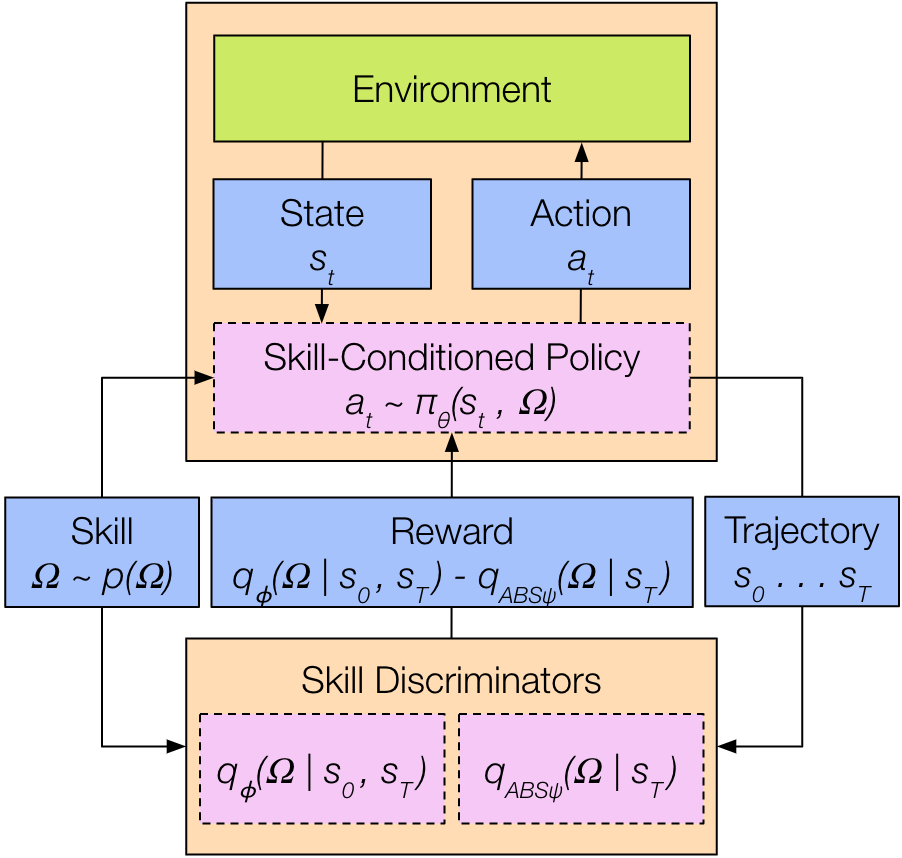}
		\caption{Relative VIC: A skill-conditioned policy interacts with
		the environment yielding a sampled trajectory. Two skill predictors are
		trained to predict the skill from the first and last states of the
		trajectory and only the last state in the trajectory respectively. The difference
		between the probability assigned by the predictors to the skill used to
		generate the trajectory is given to the policy as reward to incentivize
		learning a set of skills that is diverse in how each skill changes the
		agent's relationship to the environment.}
    \label{fig:rvic-diagram}
\end{figure}

\section{Relative Variational Intrinsic Control}

A drawback of the VIC approach is that, given sufficient trajectory lengths, it tends to yield skills which partition the state space with respect to their terminal states.
When this occurs, the inverse predictor $q$ learns to ignore the initial state $s_0$ and infers skills based only on the final state $s_T$ while the skill-conditioned policy learns to go to a different unique state for each skill $\Omega$.

This can be explained by noting that the entropy of the skill given the final state is an upper bound on the entropy of the skill given the full trajectory:
\begin{equation}
	0 \leq H(\Omega|s_T,s_0) \leq H(\Omega|s_T),
\end{equation}
so conditioning on the initial state will not increase the mutual information if the skills are perfectly predictable from just the final state.
In large or infinite environments, there are no shortage of diverse (yet potentially meaningless) final states from which to perfectly predict skills, driving the VIC objective to collapse into state-space partitioning.

Inspired by this observation, we introduce a second inverse predictor, $q^\mathrm{abs}_\psi(\Omega|s_T)$, which is trained to predict a skill's identity from the final state of the trajectory alone.
We dub this predictor $q^\mathrm{abs}$ because it can only base its predictions
on the ``absolute'' state of the environment upon skill termination, while $q$ can make use of the agent's initial state in order to utilize information about the final state \textit{relative} to the initial state.
We then train the policy \textit{adversarially} with respect to this secondary objective: we reward discriminability by $q$ while simultaneously punishing discriminability by $q^\mathrm{abs}$.

Minimizing the difference of these predictors also has its own information theoretic interpretation.
If both predictors perfectly estimate their respective conditional distributions, then the optimization process amounts to maximizing the mutual information between the skills and the initial state, given the final state:
\begin{align}
	I(s_0, \Omega|s_T)  &= H(\Omega|s_T) - H(\Omega|s_T,s_0) \nonumber \\
	&= \mathbb{E}_\Omega\mathbb{E}_{\pi_\Omega} \log p(\Omega|s_T, s_0)-\log p(\Omega|s_T) \nonumber \\	
	&\approx \mathbb{E}_\Omega\mathbb{E}_{\pi_\Omega} \log q_\phi(\Omega|s_T, s_0) - \log q^\mathrm{abs}_\psi(\Omega|s_T). \nonumber \\	
\end{align}
Since the skills are sampled independently from the initial state, maximizing this mutual information implies that each skill must communicate information about the initial state through its policy. This reinforces our intuition that our adversarial predictors should yield \textit{relative} skills. 

Note that we are not maximizing a lower bound on this mutual information, as minimizing the discriminability of $q^\mathrm{abs}$ \textit{upper} bounds $H(\Omega|s_T)$.
Previous work has shown this to be unproblematic \cite{sharma2019dynamics}, likely due to these conditional distributions being relatively easy to accurately approximate with modern over-parametrized models.

In the experiments that follow, we instantiate $q$ and $q^\mathrm{abs}$ as neural networks which share parameters, specifically a convolutional sub-network which processes pixel observations.
The observations $s_0$ and $s_T$ are each processed independently with this sub-network.
The resulting representation of $s_T$ is then processed by a multi-layer perceptron whose parameters are specific to $q^\mathrm{abs}$, while the concatenated representations of $s_0$ and $s_T$ are processed by another multi-layer perceptron representing $q$.

For the policy, we train R2D2~\citep{kapturowski2018recurrent} on fixed length ``skill episodes'' constructed on top of environment-specified episodes such that the final observation of one skill episode becomes the initial observation of the next, with each actor periodically resetting the base environment, following \citet{warde2018unsupervised}.
While \citet{gregor16variational} employed rewards in the log domain, we find that a difference of probabilities $q_\phi(\Omega|s_T, s_0) - q_\psi^\mathrm{abs}(\Omega|s_T)$ works well in practice.

\newcommand{\State}{s}
\newcommand{\Action}{a}
\newcommand{\Reward}{r}
\newcommand{\EpisodeLength}{T}
\newcommand{\ActionSpace}{\mathcal{A}}
\SetKwInOut{Input}{Input}
\SetKwInOut{Output}{Output}
\newcommand{\PolicyParams}{\theta}
\newcommand{\Policy}{\pi_\PolicyParams}
\newcommand{\Option}{\Omega}
\newcommand{\OptionDistribution}{p(\Option)}
\newcommand{\Predictor}{q}
\newcommand{\RelParams}{\phi}
\newcommand{\AbsParams}{\psi}
\newcommand{\RelPredictor}{\Predictor_\RelParams}
\newcommand{\AbsPredictor}{\Predictor^{\mathrm{abs}}_\AbsParams}
\SetKwBlock{With}{with}{}
\SetKwBlock{Otherwise}{otherwise}{}
\SetCommentSty{textrm}
\begin{algorithm}[t]
\DontPrintSemicolon
\Input{
Environment dynamics $p_E$,
	behavior policy $\Policy$,
	policy parameters $\PolicyParams$,
	relative predictor parameters $\RelParams$,
	absolute predictor parameters $\AbsParams$,
	skill episode length $\EpisodeLength$,
	discount $\gamma$,
	final step discount $\gamma_T$,
	skill episode count $M$.
}
\Repeat{termination}{
	$\State_0 \sim p(\State_0)$ \tcc*[r]{Reset the environment}
	\For{$m \leftarrow 1\ldots M$}{
	$\Option \sim \OptionDistribution$ \;
	\For{$t \leftarrow 1\ldots\EpisodeLength$}{
		Observe state $\State_{t-1}$ \;
		$\Action_t \sim \Policy(\Action|\State_{t-1}, \Option)$ \;
		$\State_{t} \sim p_E(\State_{t} | \State_{t-1}, \Action_t)$ \;
	}
	\tcc{Give same reward, post-hoc, for all steps}
	$ \Reward_{1}^\EpisodeLength \leftarrow \RelPredictor(\Option|\State_\EpisodeLength, \State_0) - \AbsPredictor(\Option|\State_\EpisodeLength) $ \;
	$\gamma_1^{\EpisodeLength - 1} \leftarrow \gamma$ \tcc*[r]{$\gamma_T$ given as separate input}
	Update $\PolicyParams$ with an off-policy reinforcement learning algorithm on $(\Action_1^\EpisodeLength, \State_0^\EpisodeLength, \Reward_1^T, \gamma_1^\EpisodeLength)$,  \;
	Update $\RelParams$ by ascending $\nabla_\RelParams \log \RelPredictor(\Option | \State_\EpisodeLength, \State_0)$ \; 
	Update $\AbsParams$ by ascending $\nabla_\AbsParams \log \AbsPredictor(\Option | \State_\EpisodeLength)$ \;
	\lIf{$m < M$}{$\State_0 \leftarrow \State_T$} }
}
\caption{\textsc{Relative VIC} \label{algo}}
\end{algorithm}
 See Figure \ref{fig:rvic-diagram} and Algorithm \ref{algo} for further summary of the Relative Variational Intrinsic Control method.  
 \section{Experiments}
In this section, we evaluate the skills learned by RVIC both qualitatively and quantitatively (via hierarchical reinforcement learning) on the DeepMind Control Suite \cite{tassa2018deepmind} and Atari 2600 games from The Arcade Learning Environment (ALE) \cite{bellemare2013arcade}.
All experiments on both Atari and DeepMind Control Suite domains are done from pixels. 
On experiments on the DeepMind Control Suite, we first discretize the continuous action space to enable value learning with R2D2. 
Using a fixed discount at every time step and allowing bootstrapping between skill episodes worked best for experiments on the DeepMind Control Suite, while experiments on Atari performed better when given a zero discount at the end of each skill episode, in addition to a zero discount upon loss of life.
All final values used for hyperparameters can be found in Table \ref{tab:table1} in the Appendix.

\begin{figure*}[ht!]
    \centering
    \includegraphics{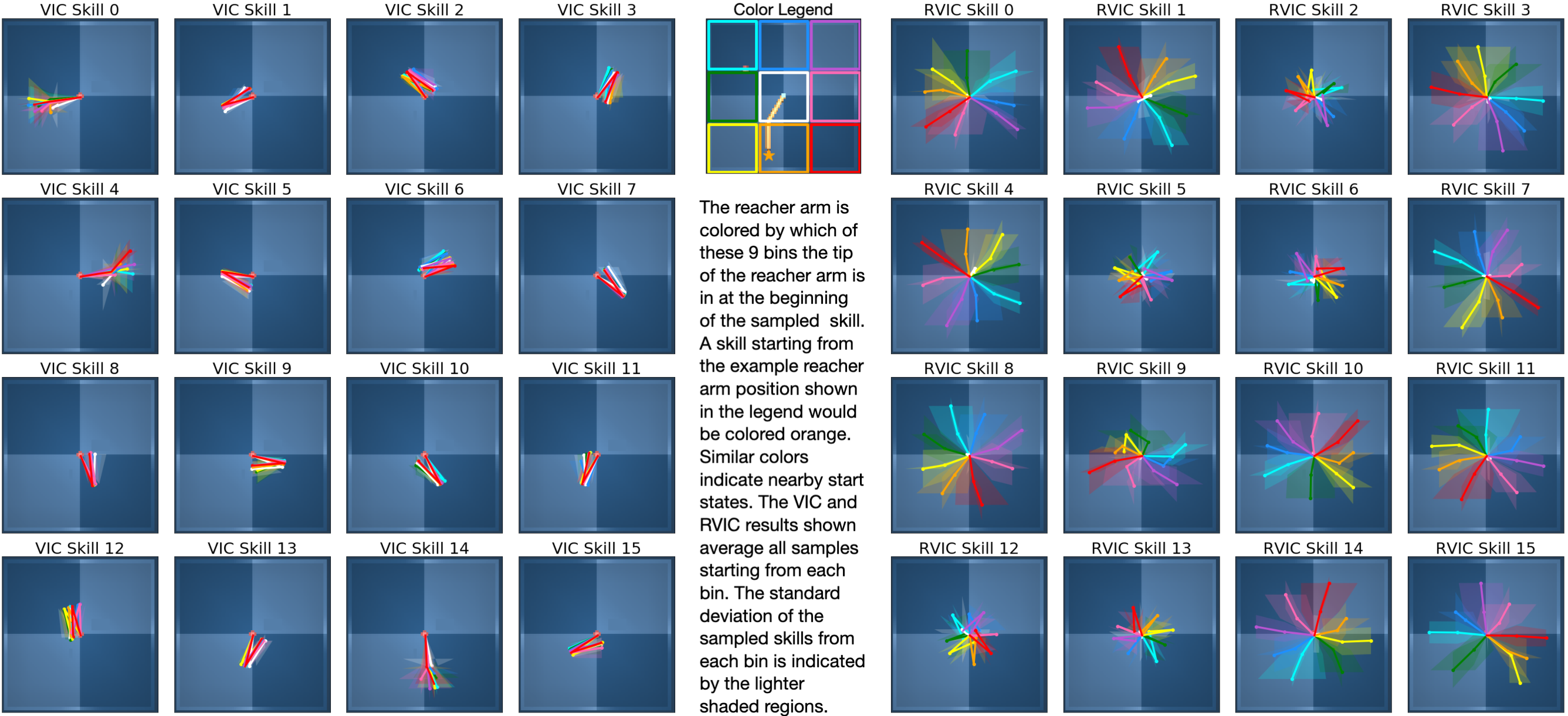}
		\caption{VIC (left) and RVIC (right) skills on Reacher in the DeepMind Control Suite.
		Each facet shows the end state of the reacher arm for a skill overlaid across multiple runs with different start states (indicated by color - see legend for details).
	}
    \label{fig:reacher-skills}
\end{figure*}
As a baseline for all experiments, we compare against VIC \cite{gregor16variational} with a fixed skill prior (as demonstrated to work better in DIAYN \cite{eysenbach2018diversity}).
DIAYN is explicitly shown in the analysis of \citet{eysenbach2018diversity} to learn skills that partition the state-space, though we do not compare against DIAYN directly, as it does not attempt to condition the inverse predictor on the initial state, instead conditioning it on any independently drawn sample state from the trajectory. 
Therefore, VIC, which tries (but often fails) to condition the inverse predictor on the initial state is a more relevant baseline to compare against.
We do not compare against \citet{sharma2019dynamics} as the method is constrained to working with crafted features rather than pixels, and it is non-obvious how to adapt DADS to work for experiments from pixels.
Our choice of baseline is therefore the closest ablation to Relative VIC, as the only major difference between the two methods is the two-predictor reward objective used by RVIC.

For both methods, we experimented with giving skill rewards in dense way, where the reward calculated for the entire skill episode $q_\phi(\Omega | s_0, s_T) -q^\mathrm{abs}_\psi(\Omega | s_T)$ is given at every timestep t in the skill trajectory, which is easily done in an off-policy learning set up.
We found this dense reward to work better empirically for both methods than the sparse reward used in VIC.
Skills from both methods were trained for 175 million learner steps before being analyzed qualitatively and used in HRL experiments.

\begin{figure*}
    \centering
    \includegraphics{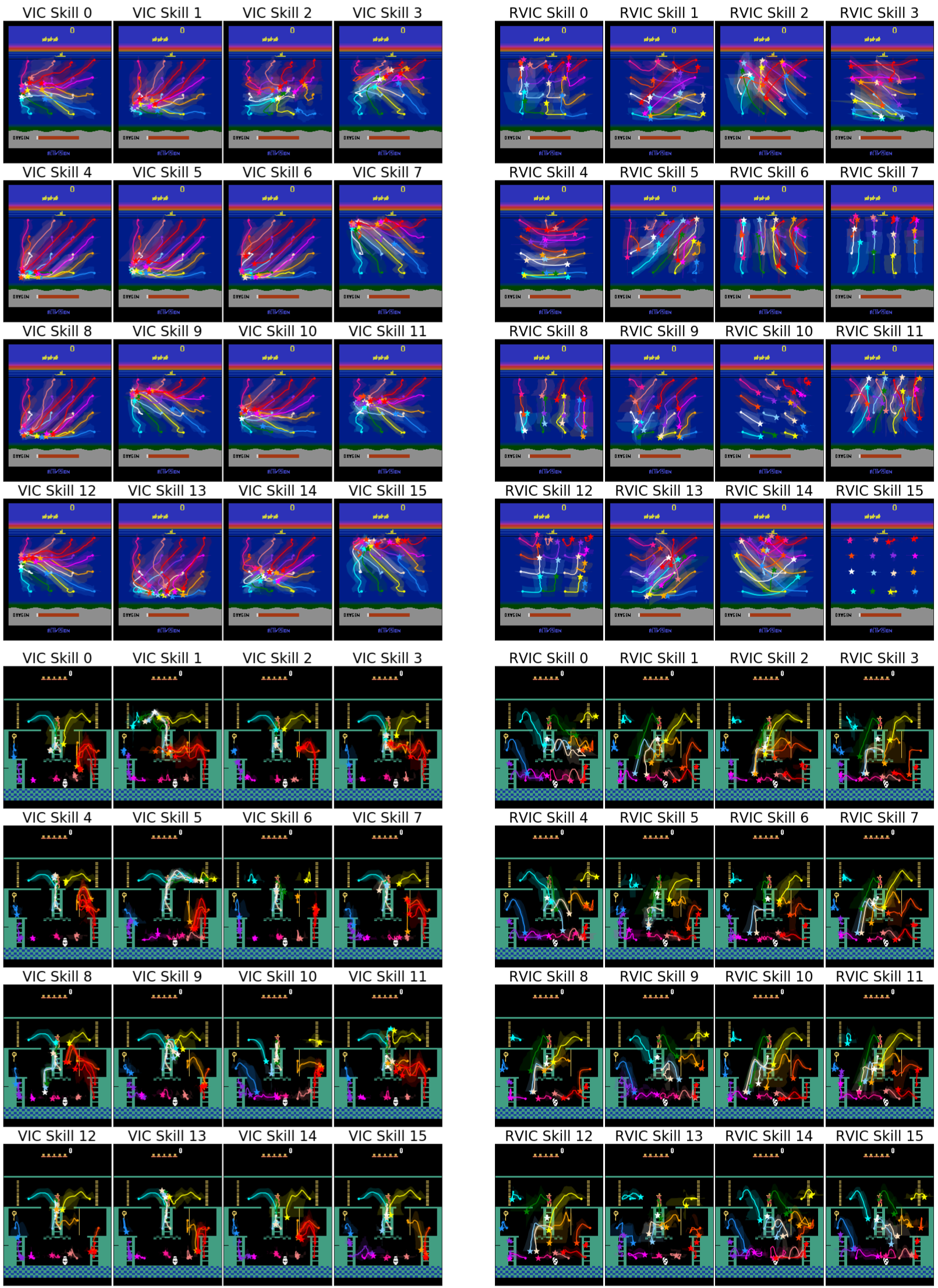}
		\caption{Learned VIC (left) and RVIC (right) skills on Seaquest (top) and Montezuma's Revenge (bottom).}
    \label{fig:atari-skills}
\end{figure*}

\begin{figure*}[ht!]
    \centering
    \includegraphics[width=1.0\textwidth]{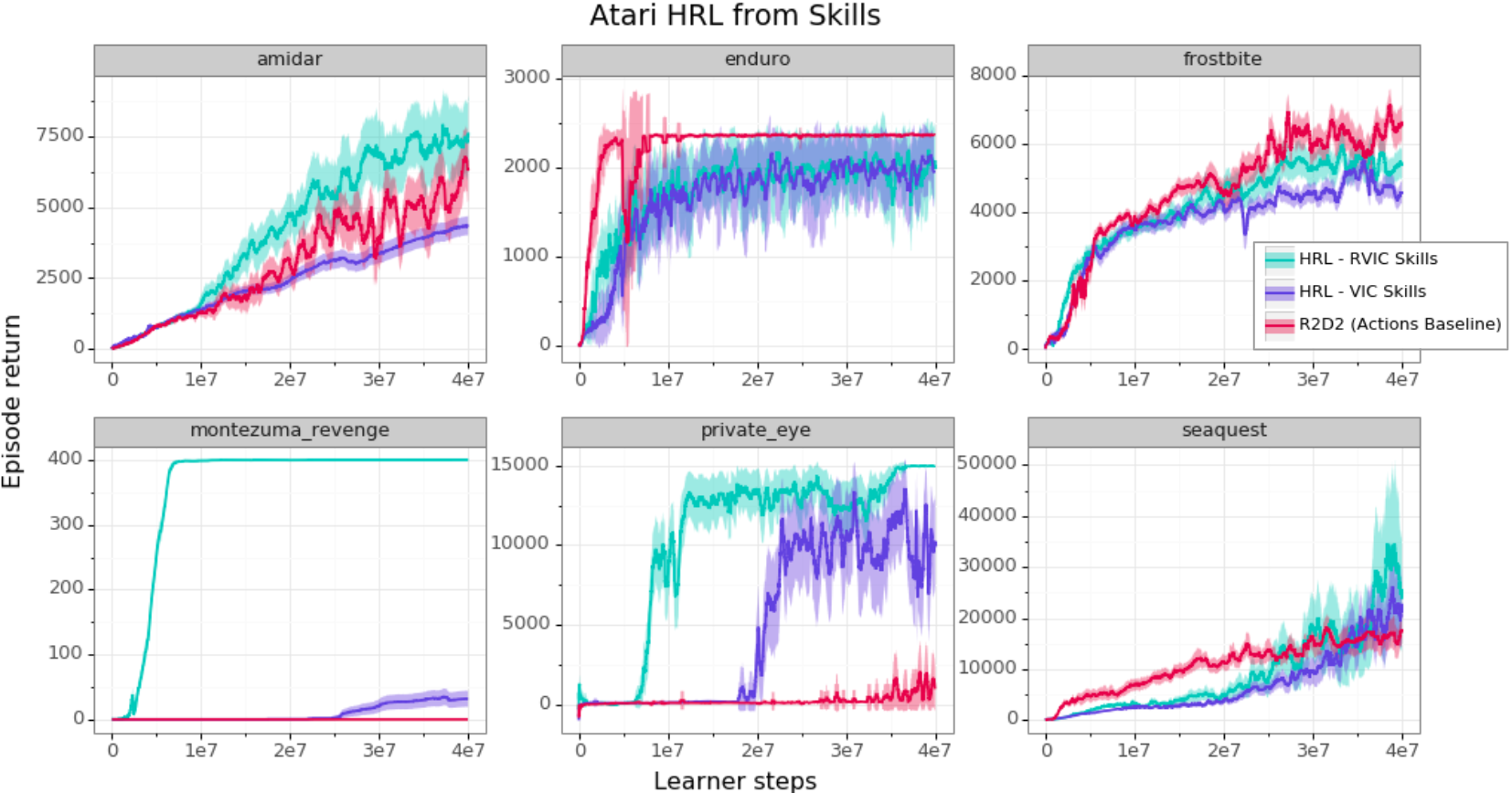}
	  \caption{HRL Results.
	           The option to execute a pre-trained skill for a fixed duration is added to the set of possible actions.}
    \label{fig:hrl-results}
\end{figure*}

\subsection{Qualitative Results}
For the qualitative experiments, both methods learn a set of 16 skills with a skill episode length of 90 for experiments on the DeepMind Control Suite and 25 for experiments on Atari.
We want to observe how each skill behaves from various start states.
We visualize how skills learned by the two skill learning methods affect the controllable environment state.
In the Reacher domain on the DeepMind Control Suite, the controllable part of the environment is the position of the two-link reacher arm.
Visualizing how skills from each method change the position of the reacher arm from various starting positions yields insights into the difference between baseline VIC and RVIC.

While the baseline VIC skills on the left in Figure \ref{fig:reacher-skills} reliably reach diverse end states, the skills disregard their starting positions (indicated by color) in achieving that diversity.
Regardless of where the reacher arm starts, each skill only goes to a single end state.
Additionally, note how the baseline VIC skill diversity only spans a portion of the state space and the reacher arm is rarely even opened as it is easy enough to achieve the VIC objective without doing so.
In contrast, the set of RVIC skills shown partitions the rotation space of both joint angles, therefore covering more of the relevant state space.
In other words, an RVIC skill will rotate the joint angles a certain amount from wherever the reacher arm starts. 
This can be seen in the right side of Figure \ref{fig:reacher-skills} as the ordering of colors (where nearby colors indicate nearby start states) is preserved in each skill, but rotated to a different degree for each skill.

Similarly, on Atari games Seaquest and Montezuma’s Revenge, we visualize how different skills move the avatar through the X-Y plane, since the agent has direct control over only the (X, Y) coordinates of the avatar.
We extract this information about the avatar coordinates from RAM only for the visualizations, as all learning is done directly from pixels.
To clearly show how each skill learning method utilizes (or disregards) information about the start state, we roll out sample skills from every possible starting (X, Y) coordinate in the
frame (excluding positions in Montezuma's Revenge where the avatar is free-falling through the air and has no control over its behavior).
The samples are then divided into 16 bins (uniform 4x4 grid for Seaquest; meaningfully different state areas such as platforms or ladders in Montezuma's Revenge).
Sample trajectories from each bin are then averaged, colored such that similar colors indicate nearby bins, and plotted along with their standard deviation in Figure \ref{fig:atari-skills}.
The final state of each averaged trajectory is denoted with a star. 

The baseline VIC method on Seaquest obtains skill diversity via state-space partitioning, converging to a single final state no matter where the trajectory starts out.
Partial state-space partitioning behavior of VIC can also be seen to a lesser degree on Montezuma's Revenge. 
Due to the difficulty in Montezuma's Revenge of exploring to the bottom parts of the frame during unsupervised training, the VIC skills seem to only pay attention to partitioning the top part of the frame with some skills clearly choosing a target state, for example, the top of the center ladder in skill 13. 
When the avatar coordinates are set to the bottom of the frame at analysis time, the agent has probably never seen those states before and likely does not know how to reach the skill's target states.
When performing these VIC skills from the unfamiliar positions at the bottom of the frame, behaviors are inconsistent with the behaviors learned for the top of the frame, often doing nothing at all. 
This illustrates the dangers of skills that partition the state space as behavior does not generalize well to the unseen states at the bottom of the frame and skills do not perform consistent behavior everywhere.
The skills learned by RVIC on both of these games are different depending on where the agent starts out and can loosely be seen as directional skills that are consistent, generalizable, and composable.

\subsection{Hierarchical RL on Skills}
Hierarchical Reinforcement Learning (HRL) decomposes the RL problem into temporally extended sub-problems solved with multiple hierarchical levels of planning or control \cite{sutton1999between, precup2001temporal}.
To test the usefulness of the learned skills, we perform HRL experiments on six Atari games, using the pre-trained skill policies as the low-level controller.
After training skills for both the baseline and RVIC, we introduce a second phase of training where a meta-controller agent can use the pre-trained, frozen skill policies to maximize the extrinsic rewards from the environment.
In this second phase, a meta-controller learns a policy that at each timestep can either act with a temporally-extended skill or a primitive action for a single timestep.
From the perspective of the meta-controller, a skill is just another type of action available, though what actually happens after acting with a skill is temporally extended behavior. 
Rewards collected during the skill episode are appropriately summed and discounted over the skill duration and returned to the meta-controller policy. 

Since at the beginning of training the meta-controller may be incentivized to
only select primitive actions to obtain more fine-grained control, we also
introduce a fixed meta-action cost (similar to the "deliberation cost" in \citet{harb2018when}) that is deducted from the reward given to the agent every time the meta-controller selects an action, incentivizing it to use the temporally extended skills where possible.
Like the skill-policies, the meta-controller policy is also trained using R2D2, though prioritized experience replay is disabled as priorities are inconsistent to calculate between single-step primitive actions and temporally-extended behaviors.
Additionally, we use shorter unroll length (40) and burn-in length (10) than plain R2D2 to account for the use of temporally-extended behaviors.

For both skill HRL variants, we choose the best meta-action cost from (0.0, 0.1, 0.15) and the best skill episode length from (10, 15, 25) for each game. The
best values for each Atari level are recorded in Table \ref{tab:table2} in the Appendix.
In all levels, the number of skills is fixed to 16.
Results are averaged over three random seeds. 

In Figure \ref{fig:hrl-results}, we compare the speed of learning over the first 40 million learner steps of an R2D2 agent that acts with either Relative VIC skills and primitive actions, baseline VIC skills and primitive actions, or just primitive actions.
The results are mixed across levels although the Relative VIC skills show a learning advantage over the baseline VIC skills and action-baseline R2D2 on most of the games.

To explain the advantage of RVIC skills, we note that in an environment that is sufficiently large or hard to explore during unsupervised training (for example, Montezuma’s Revenge which has many rooms that are hard to find by accident), a skill set that partitions the state space will likely restrict skills to partition an incomplete part of the state space, (say, only
the first room in Montezuma’s Revenge).
Using such a set would never allow a meta-controller to explore outside of the already seen subset of the state-space.
RVIC can be seen as removing potential overfitting to the distribution of final states seen during the unsupervised pre-training of skills.

 \section{Related Work}
The idea of maximizing the mutual information between an agent's behavior and the outcome of that behavior can be traced back to \citet{klyubin2005empowerment}, where the term `empowerment' was coined to describe this objective.
The empowerment objective was initially limited to small domains due to the costliness of mutual information estimation.
\citet{mohamed2015variational} introduced the idea of maximizing a variational lower bound to scale the empowerment objective to larger domains.
Specifically, the Barber-Agakov bound was used \citep{barber2003algorithm}, which decomposes the mutual information into a difference of entropies.
The negative condition entropy can be lower-bounded by learning a `reverse predictor'.
This can then be combined with knowledge of the true marginal entropy term (e.g. the policy) to provide a lower bound on the mutual information.
\citet{gregor16variational} combined variational empowerment with a latent variable for capturing closed-loop temporally extended behavior (i.e. an `option', or, as we refer to in this paper, a 'skill').
This approach, Varitional Intrinsic Control (VIC), parametrized the skill distribution, policy, and skill predictor with neural networks.
A skill was sampled and used to condition the policy for some duration.
Subsequently, the reverse predictor would predict the sampled skill from the initial and final state.
The entropy of the skill distribution and the skill prediction were optimized directly, and reward functions were derived for appropriate credit assignment to the skill distribution and policy.

`Diversity is All You Need' (DIAYN) simplified the VIC algorithm by fixing the option distribution to be the marginal maximum entropy distribution, which most subsequent methods have done as well, including all of those presented here \citep{eysenbach2018diversity}.
While this work also added an action entropy term to the objective, we follow \citet{hansen2019fast} in disregarding it, since it is generally less beneficial in discrete-action environments like the Atari suite.
Indeed, DIAYN differs significantly from this work in that it only considered environments with explicit state-representations.
 This both simplifies the perception aspect of the policy learning problem as well as provides a strong inductive bias to the reverse predictor by way of forcing predictability to only arise from these dimensions.
\citet{hansen2019fast} does show strong results on pixel-based environments, but only when using continuous skill distributions that would significantly increase the burden (by exploding the effective action space) when used for down-stream hierarchical reinforcement learning.

`Discriminative Embedding Reward Networks' (DISCERN) introduced the idea of chaining together sampled skills rather than resetting the environment state between each one \citep{warde2018unsupervised}.
This greatly increases the entropy of the initial state distribution, which increases the difficulty of the learning problem in exchange for decoupling skill duration from the final state distribution.
For example, if skills reset the environment on termination, then the skill duration would have to be hundreds of steps long to get anywhere in most Atari games.
Our method differs from DISCERN in that we are explicitly interested in relative skills as opposed to the achievement of absolute goals represented by desired observations.

\citet{achiam2018variational} propose a trajectory-conditional reverse predictor motivated by the idea of learning diverse `behaviors' rather than diverse goals. Like DIAYN, this method has not been shown to be effective on pixel-based environments and relies on a low-frequency heuristic that would likely be inappropriate for learning skills in Atari games.

Finally, \citet{sharma2019dynamics} maximize a partial lower bound in the same sense that RVIC does.
Namely, a difference of entropies decomposition is used even though the marginal entropy is not known a priori and must also be approximated.
However, this method is also only shown to work from explicit state-representations and it is non-obvious how to modify it to work from pixels.
The empirical stability of both methods suggest that a `proper' lower bound on a mutual information is not necessary for empowerment based approaches to succeed.

The concept of affordances was first introduced by \citet{gibson1977theory} within psychology to refer to action possibilities or opportunities the environment affords an animal at any given time.
An affordance emerges from the relationship between an agent and its environment.
\citeauthor{gibson1977theory} suggested that humans are able to easily perceive affordances.
Further, \citeauthor{gibson1977theory} argues that humans actively alter the environment to change what the environment affords us.
Within the context of reinforcement learning, \citet{cruz2014improving} demonstrate that giving agents affordances as prior knowledge can greatly speed up convergence, and \citet{khetarpal2020can} demonstrate the ability to use affordances to aid planning in RL with partial-models and enable better generalization.

Our work differs from these approaches in its ability to learn affordance-like skills directly from interaction, without any prior knowledge as to what kinds of behavior the environment affords.
Indeed, RVIC can be seen as a possible mechanism by which knowledge of affordances can arise.
Integrating this mechanism with the rest of the literature on affordances is a promising avenue for future work. 
 \section{Discussion}
Relative VIC learns meaningfully diverse skills that partition the space of affordances by incentivizing the skills to be distinguishable given their first and last states but not distinguishable given only the last state.
We analyzed the difference between Relative VIC learned skills and VIC learned skills on several domains and demonstrate the ability to learn affordance-like skills from pixels in both the DeepMind Control Suite and Atari domains.
We demonstrate the usefulness of Relative VIC skills in the Hierarchical RL framework on Atari and their ability to generalize across various parts of the state space.

Some limitations of the method include the use of a fixed discrete uniform skill prior which implies that all skills should exist at every state, even if some options don't make sense at every state.
Additionally, the fixed skill duration may prove to be too rigid for some environments to use effectively in HRL.
Other potential future direction of this work include learning the meta-controller policy and skill policies simultaneously.
 
\bibliography{paper.bib}
\newpage
\appendix
\appendixpage
\section{Hyperparameters}
\begin{table}[h!]
  \begin{center}
    \begin{tabular}{|l|c|}
			\hline
      Number of actors & 256\\
      Batch size& 64\\
			Optimizer & Adam \shortcite{kingma2014adam}\\
			Dense skill reward & True\\
      Skill episode count & 10\\
      $q_\phi$ learning rate & $10^{-4}$\\
			$q_\psi^{abs}$ learning rate & $10^{-4}$\\
      $\pi_\theta$ learning rate& $10^{-4}$\\
			$\pi_\theta$ target update period & 10000\\
			$q_\phi$, $q_\psi^{abs}$ target update period & 10\\
			Actor update period & 100\\
			0 $\gamma$ at skill end (Control Suite) & False\\
			0 $\gamma$ at skill end (Atari) & True\\
			Skill length (Control Suite) & 90\\
			Skill length (Atari) & See Table 2\\
			Number of skills & 16\\
			$q_\phi$, $q_\psi^{abs}$ shared torso & DQN Conv Torso\\
			$q_\phi$, $q_\psi^{abs}$ head hidden size & 512\\
			\hline
    \end{tabular}
    \caption{A table of the final hyperparameters for skill learning experiments.
		Hyperparameter values are shared across both ALE Atari and DeepMind Control Suite domains
		and RVIC/VIC experiments unless otherwise specified. Network architecture
		parameters for training both the skill-conditioned policies and the HRL
		meta-controller policies are taken from the R2D2 paper unless otherwise
		stated.}
    \label{tab:table1}
  \end{center}
\end{table}
\begin{table}[h]
  \begin{center}
		\begin{tabular}{|l|c|c|c|c|}
			\hline
			& \multicolumn{2}{c}{\textbf{Baseline VIC}} & \multicolumn{2}{|c|}{\textbf{Relative VIC}}\\
			\cline{2-5}
			& Meta & Skill & Meta & Skill\\
			& action & episode& action & episode\\
			& cost & length & cost & length\\
			\hline
			\textbf{Amidar} & 0.15 & 25 & 0.15 & 10\\
			\textbf{Enduro} & 0.00 & 10 & 0.00 & 10\\
			\textbf{Frostbite} & 0.15 & 10 & 0.00 & 10\\
			\textbf{M. Revenge} & 0.00 & 25 & 0.10 & 10\\
			\textbf{Private Eye} & 0.00 & 10 & 0.15 & 25\\
			\textbf{Seaquest} & 0.15 & 10 & 0.00 & 10\\
			\hline
    \end{tabular}
    \caption{A table of the best hyperparameters chosen for each level of Atari for HRL experiments.
		The hyperparameter combination with the best HRL performance when averaged over 3 random seeds was chosen and results are displayed in Figure \ref{fig:hrl-results}.}
    \label{tab:table2}
  \end{center}
\end{table}

\end{document}